\documentclass[11pt,a4paper]{article}
\usepackage[hyperref]{acl2020}
\usepackage{times}
\usepackage{latexsym}
\usepackage{amssymb}
\usepackage{amsmath}
\usepackage{booktabs}
\usepackage{multirow}
\usepackage{multicol}
\renewcommand{\footnotesize}{\tiny}
\usepackage{tipa} 

\newcommand{\imagechat}{Image-Chat}


\usepackage[]{todonotes}

\newcommand*\iftodonotes{\if@todonotes@disabled\expandafter\@secondoftwo\else\expandafter\@firstoftwo\fi}

\usepackage{microtype}

\aclfinalcopy 

\title{
Image-Chat: Engaging Grounded Conversations
}

\author{Kurt Shuster, Samuel Humeau, Antoine Bordes, Jason Weston\\
  Facebook AI Research \\
  \texttt{\{kshuster,samuelhumeau,abordes,jase\}@fb.com}}
\date{}

\begin{document}

\maketitle
\begin{abstract}
To achieve the long-term goal of machines being able to engage humans in conversation,
our models should captivate the interest of their speaking partners. 
Communication grounded in images, whereby a dialogue is conducted based on a given photo,
is a setup
naturally appealing
to humans \citep{hu2014we}.  In this work we study large-scale architectures
and datasets for this goal. 
We test a set of neural architectures using state-of-the-art 
image and text representations, 
considering various ways to fuse the components.
To test such models, we collect a dataset of grounded human-human conversations,
where speakers are asked to play roles given a provided emotional mood or style, 
as the use of 
such traits
is also a key factor in engagingness \cite{guo2019mscap}. 
Our dataset, Image-Chat, consists of 202k dialogues 
over 202k images
using 215 possible style traits.
Automatic metrics and human evaluations of engagingness show the efficacy of our approach; 
in particular, we obtain state-of-the-art performance on the existing IGC task,
and our best performing  model is
almost on par with humans 
on the Image-Chat test set (preferred 47.7\% of the time). 
\end{abstract}

\section{Introduction}
A key way for machines to exhibit intelligence is for them to be able to
perceive the world around them -- and to be able to communicate with humans in natural language about that world.  To speak naturally with humans it is necessary to understand the natural things that humans say about the world they live in, and to respond in kind. This involves understanding what
they perceive, e.g. the images they see, what those images mean semantically for humans, and how mood and style  shapes the language and conversations derived from these observations. 

In this work we take a step towards these goals by considering grounded dialogue 
involving open-ended discussion of a given image, a setting that is naturally fun for humans \citep{hu2014we},
and study neural conversational models  for task. 
In particular, 
we explore both generative and retrieval models that
handle multimodal dialogue by fusing
Transformer  architectures \cite{vaswani2017attention}
for encoding dialogue history and responses and ResNet architectures \cite{he2016deep}
for encoding images. 
We propose ways to fuse those modalities together and perform a detailed study  including both automatic evaluations, ablations and human evaluations of our models using crowdworkers.

To train and evaluate such models, 
we collect a large set of human-human crowdworker conversations, with the aim of training a model to engage a human in a similar fashion, consisting of 202k
diverse images and 401k utterances over the images, with 215 different  style traits (e.g., optimistic, skeptical or frivolous) to promote engaging conversation.
The dataset is made publicly available in ParlAI \cite{miller2017parlai} \footnote{\tiny\url{http://parl.ai/projects/image_chat}}.

Our results show that there is a significant gap between state-of-the-art 
retrieval and generative models on this task. Our best fused retrieval 
models set a strong baseline, 
being preferred to human conversationalists  47.7\% of the time.
We show that both large-scale image and text pre-training, and utilization of style traits, are critical for best results.
We then consider transfer to the existing Image Grounded Conversations (IGC) 
task of \citet{mostafazadeh2017image}, where we obtain state-of-the-art results.

\section{Related Work}
The majority of work in dialogue is not grounded in perception, e.g. much recent work explores sequence-to-sequence models or retrieval models for goal-directed \citep{henderson2014second}
or chit-chat tasks \citep{vinyals2015neural,zhang2018personalizing}.
While these tasks are text-based only, many of the techniques developed can likely be transferred for use in multimodal systems, for example using state-of-the-art Transformer representations for text \cite{mazare-etal-2018-training} as a sub-component.

In the area of language and vision, one of the most widely studied areas is image captioning,
whereby a single  utterance is output given an input image. This typically involves producing a factual, descriptive sentence describing the image, in contrast to producing a conversational utterance as in dialogue. Popular datasets
include  COCO \citep{chen2015microsoft} and Flickr30k \citep{young2014image}. Again,
a variety of
sequence-to-sequence  \citep{vinyals2015show,xu2015show,anderson2017bottom} and retrieval models  
\citep{8578848,faghri2018vse++,Nam2016DualAN} have been applied. These tasks measure the ability of models to understand the content of an image, but not to carry out an engaging conversation grounded in perception. Some works have extended image captioning from being purely factual towards more engaging captions by incorporating style  while still being single turn, e.g. \citep{mathews2018semstyle,mathews2016senticap,gan2017stylenet,guo2019mscap,Shuster_2019_CVPR}. 
Our work also applies a style component, but concentrates on image-grounded dialogue, rather than image captioning.

\begin{figure*}[!tbp]
\centering\setlength{\tabcolsep}{0.5em}
\begin{tabular}{p{12.75em}p{12.75em}p{12.75em}}
 & & \\[-0.62em]
\includegraphics[width=12.75em, height=7.8em]{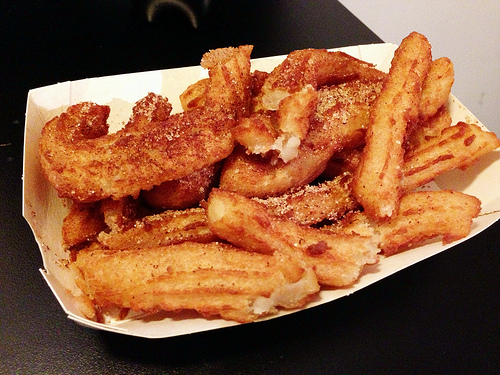} & \includegraphics[width=12.75em, height=7.8em]{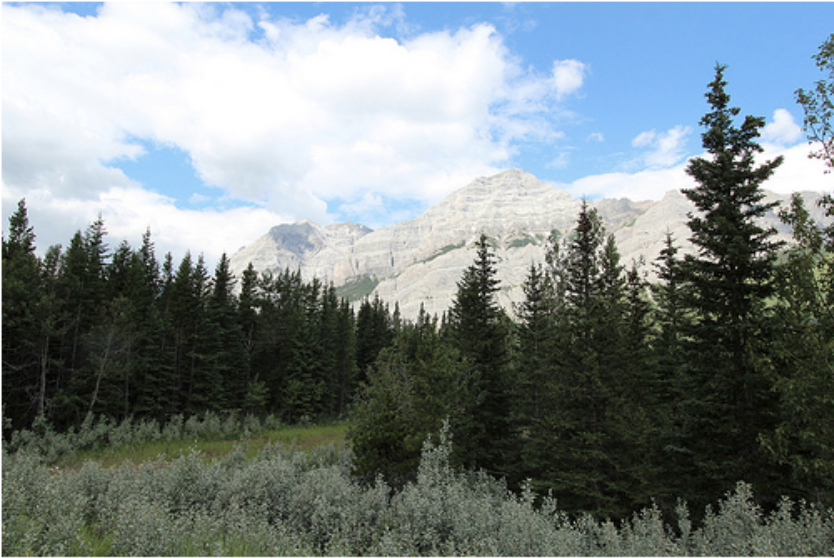} & \includegraphics[width=12.75em, height=7.8em]{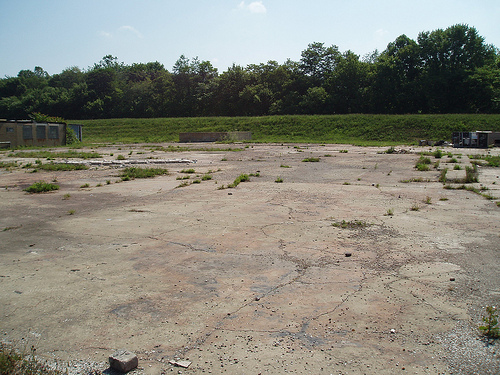} \\
\textit{\small{A: Peaceful~~~~~~B: Absentminded }} & 
\textit{\small{A: Fearful~~~~~~~B: Miserable}} & 
\textit{\small{A: Erratic~~~~~~~B: Skeptical}} \\ 
\small{A: I'm so thankful for this delicious food.} 
& \small{A: I just heard something out there and I have no idea what it was.}
& \small{A: What is the difference between the forest and the trees? Oh look, dry pavement.} \\ 
\small{B: What is it called again?} 
& \small{B: It was probably a Wolf coming to eat us because you talk too much.}
& \small{B: I doubt that's even a forest, it looks like a line of trees.} \\ 
\small{A: Not sure but fried goodness.} 
& \small{A: I would never go camping in the woods for this very reason.}
& \small{A: There's probably more lame pavement on the other side!} \\ 
 & & \\[-0.62em]
\end{tabular}
\caption{Some samples from the \textsc{\imagechat} training set. For each sample we asked humans to engage in a conversation about the given image, where the two speakers, A and B, each have a given provided style.
\label{fig:examples}
}
\end{figure*}

Visual question answering \citep{antol2015vqa} and visual dialogue \citep{das2017visual} are another set of tasks which employ vision and language. They
require the machine to answer factual questions about the contents of the image, either in single turn or dialogue form. They do not attempt to model natural conversation, but rather 
 assess whether the machine can perform basic perception over the image via a series of questions.
 
There are some works which directly address dialogue grounded with vision. 
The work of \citet{pasunuru-bansal-2018-game} assesses the ability to execute dialogue given video of computer  soccer games. The work of \citet{huber2018emotional} investigates the use of sentiment-based visual features and facial expressions for emotional image-based dialogue.
Perhaps the most related work to ours is \citet{mostafazadeh2017image}. Their work considers (visual context, textual context, question, response) tuples, and builds validation and test sets based on 4k eventful images called Image Grounded Conversations (IGC). No training data is provided, but instead the authors use Twitter for that in their experiments.
In contrast, we provide training, validation and testing sets over 202k images for our task (that do not overlap with IGC), and consider a general set of images and dialogues, not just events and questions plus responses. In our experiments we also show strong transfer ability of our models to  the IGC task.

While there are many ways to measure dialogue quality, human engagement is a popular metric.
Engagement itself can be measured in many ways \citep{bohus2009models,yu2016wizard} but here we 
adopt the common approach of simply asking humans which speaker they find more engaging, following
other works \citep{li2019acute,dinan2020second}.

\section{Image-Chat}
The \textsc{\imagechat} dataset is a large
collection of (image, style trait for speaker A, style trait for speaker B, dialogue between A \& B) tuples that we collected using crowd-workers, 
Each dialogue consists of consecutive turns by speaker A and B.
No particular constraints are placed on the kinds of utterance, only that we ask the speakers to both use the provided style trait,
and to respond to the given image and dialogue history {\em in an engaging way}.
The goal is not just to build a diagnostic dataset but a basis for training models that humans actually want to engage with.

\paragraph{Style Traits} 
A number of works have shown that style traits for image captioning help provide creative captions \citep{mathews2018semstyle,mathews2016senticap,gan2017stylenet,Shuster_2019_CVPR}. 
We apply that same principle to image grounded dialogue, considering a set of 
215 possible style traits, 
using an existing set from \citet{Shuster_2019_CVPR}. 
The traits are categorized into three classes:
positive (e.g., sweet, happy, eloquent, humble,
witty),
neutral (e.g., old-fashioned, skeptical, solemn, questioning) and negative (e.g., anxious, childish, critical, fickle, frivolous).
We apply these  to both speakers A and B, who will be assigned different style traits for each given conversation.  

\paragraph{Images}
The images used in our task are randomly selected from the YFCC100M Dataset\footnote{\tiny{https://multimediacommons.wordpress.com/yfcc100m-core-dataset/}} \citep{Thomee:2016:YND:2886013.2812802}.

\paragraph{Dialogue}

For each image, we pick at random two style traits,  one for speaker A and one for speaker B, and collect the dialogue using crowdworkers who are asked to both assume those roles, and to be engaging to the other speaker while doing so.   It was emphasized in the data collection instructions that the style trait describes a trait of the speaker, 
not properties of the content of the image they are discussing. 
Some examples from the training set are given in Figure \ref{fig:examples}. 

\paragraph{Data Quality}
During data collection crowd-sourcers were manually monitored, checking to ensure they were following the instructions. Poor performers were banned, with comments discarded.
A verification process was also conducted on a subset of the data, where separate annotators were asked to choose whether the utterance fit the image, style, or both, and found that 92.8\% of the time it clearly fit the image, and 83.1\%  the style, 
and 80.5\% both. Note, given that not all utterances should directly reference an image property or 
invoke the style, we do not expect 100\%.

\paragraph{Overall Dataset}

The overall dataset statistics are given in Table \ref{table:data_stats}.
This is a fairly large dialogue dataset compared to other existing 
publicly available datasets. For example,
PersonaChat \cite{zhang2018personalizing} (which is not grounded in images) consists of 162k 
utterances, 
while IGC \cite{mostafazadeh2017image} 
(grounded in images) consists of 4k  of validation and test set examples only,
compared to over 400k utterances in {\sc{\imagechat}}. 

\begin{table}[t]
\begin{center}
\small
\begin{tabular}{|l|ccc|cc|}
 \hline
Split & train & valid & test\\
\hline
Number of Images     &    186,782  &  5,000  & 9,997 \\
Number of Dialogues  &    186,782  &  5,000  & 9,997 \\
Number of Utterances &    355,862  &  15,000  & 29,991 \\
\hline
Style Types  &  215 & 215 & 215 \\
Vocabulary Size    & 46,371  & 9,561 & 13,550 \\
Tokens per Utterance   & 12.3 & 12.4 &12.4  \\
\hline
\end{tabular}
\caption{\textsc{\imagechat} dataset statistics.
\label{table:data_stats}
}
\end{center}
\end{table}

\section{Models}

\begin{figure}[!ht]
\begin{center}
\includegraphics[width=0.43\textwidth]{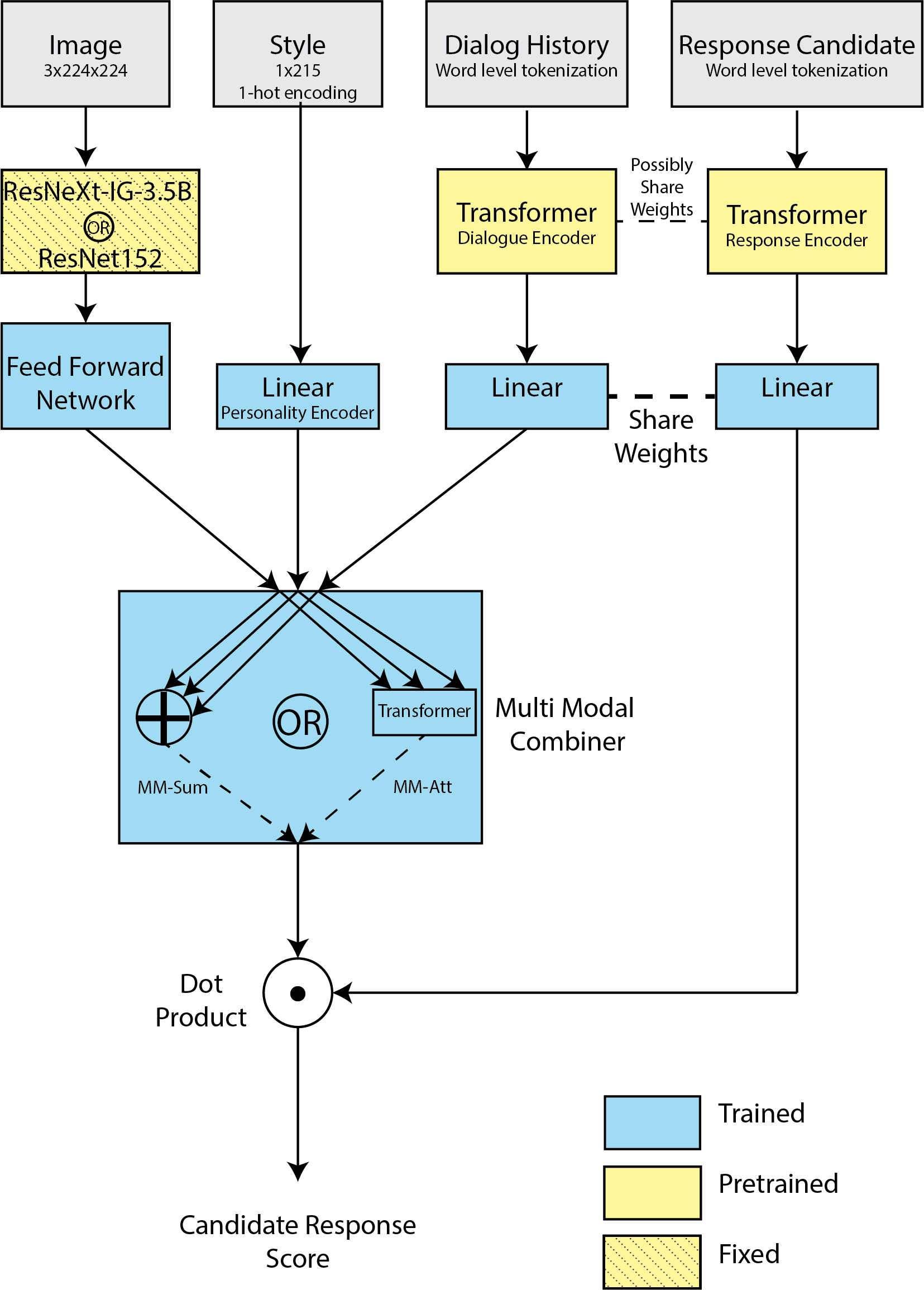}
\caption{The {\sc TransResNet$_{RET}$} multimodal architecture for grounded dialogue.  There are several options: different image encoders (ResNet152 or ResNeXt-IG-3.5B), text encoders (shared or separate Transformers for history and response), and different multimodal combiners (sum or attention-based).}
\label{transresnet_schema}
\end{center}
\end{figure}

We consider two major types of dialogue model: retrieval and generative. 
Both approaches  make use of the same components as building blocks.
 We use three sub-networks for the three modalities of input: (i) an image encoder, (ii) a dialogue history encoder; and (iii) a style encoder. 
In the retrieval model these are then fed into a combiner module for combining the three modalities. Finally, there is a response encoder for considering candidate responses and this is scored against the combined input representations. An overview of the retrieval archictecture 
is shown in Figure \ref{transresnet_schema}.
For the generative model, the three encoders are used as input,  and a further decoder Transformer is used for outputting a token sequence; beam search is applied.

\paragraph{Image Encoder}

We build our models on top of pretrained image features, and compare the performance of two types of image encoders. The first is a residual network with 152 layers described in \citet{he2016deep} trained on ImageNet \citep{imagenet} to classify images among 1000 classes, which we refer to in the rest of the paper as {\em ResNet152} features. We used the implementation provided in the torchvision project \citep{torchvision}. The second is a ResNeXt $32\times48$d \citep{resnext} trained on 3.5 billion Instagram pictures following the procedure described by \citet{uru}, which we refer to in the rest of the paper as {\em ResNeXt-IG-3.5B}. 
The representation \(r_I\) of an image \(I\) is obtained by using the 2048-dimensional output of the image encoder as input to a feed-forward network:
a multi-layer perceptron with ReLU activation units and a final layer of 500 dimensions in the retrieval case, and a linear layer in the generative case.

\paragraph{Style Encoder}  To condition on a given style trait, we embed each trait to an $N$-dimensional vector to obtain its representation \(r_S\). We used $N=500$ for retrieval 
and $N=300$ for generation.

\paragraph{Dialogue Encoder} 
The entire dialogue history $D$ is encoded into a fixed size vector \(r_D\) using a Transformer architecture \citep{vaswani2017attention}, followed by a linear layer. 
Such Transformers have been shown to perform strongly on a variety of dialogue
tasks previously \citep{yang2018learning,mazare-etal-2018-training}. 
We use a Transformer with 4 layers, 300 hidden units, and 6 attention heads. The outputs are pooled (mean) to give a final vectorial encoding.

We pretrain the entire encoder following the setup described in \citet{mazare-etal-2018-training}: we train two encoders on a next-utterance retrieval task on a Reddit dataset of dialogues containing 1.7 billion pairs of utterances, where one encodes the context and another the candidates for the next utterance; their dot product indicates the degree of match, and they are  trained with negative log-likelihood and $k$-negative sampling.
We then initialize our system using the weights of the candidate encoder only, and then train on our task in either generative or retrieval mode.

\subsection{Retrieval Models} 

\paragraph{Multimodal combiner module}
We consider two possible combiner modules for the inputs:

{{\em Multimodal sum combiner (MM-sum)}}: Given an input image, style trait and dialogue \((I, S, D)\), together with a candidate response \(C\), the score of the final combination is computed as
\(s(I, S, D, C) = (r_I + r_S + r_D) \cdot r_C\). 

{\em{Multimodal attention combiner (MM-att)}}: A more sophisticated approach is to use an attention mechanism to choose which modalities are most relevant for each example by stacking Transformers. We concatenate the three representation vectors 
$r_I$, $r_S$  and  $r_D$ and feed them to a second Transformer (4 attention heads, 2 layers, 500 hidden units) which performs self-attention over them. 
 The three modalities are thus reweighted by the corresponding attention weights to give the final input representation vector $r_{T}$, which is used to compute the score for a given candidate using 
 $r_{T} \cdot r_C$.

\paragraph{Response encoder}
We employ the same Transformer architecture as in the dialogue encoder for encoding candidate responses. We tried two variants: either sharing or not sharing the weights with the input dialogue encoder.

\paragraph{Training and Inference} 
Given a tuple \(I, S, D\), and a set of candidates \((c_1, .., c_N)\), at inference time the predicted utterance is the candidate \(c_i\) that maximizes the score \(s(I, S, D, c_i)\). At training time we pass  a set of scores through a softmax and train to maximize the log-likelihood of the correct responses. We use mini-batches of 500 training examples; for each example, we use the gold responses of the other examples of the batch as negatives. 
During final human evaluation all candidates from the training set are considered to produce a response (356k candidates in our experiments).

\subsection{Generative Models} 

\paragraph{Dialogue Decoder} 

The  encoding from the image encoder has a final linear layer of  dimension 2048 $\times$ 300. 
This projects it to the same size of the token encoding of the dialogue decoder. We thus add it as an extra token at the end of the Transformer’s encoder output. For style, we simply prepend the style to the beginning of the dialogue history, and it is thus encoded in the dialogue encoder.
We then treat this as a standard seq2seq Transformer in order to generate dialogue responses.

\paragraph{Training and Inference} 

We train with a batch size of 32 and learning rate of $.0001$ using adam, 
and apply beam search with a beam of size 2 and tri-gram blocking at 
inference time. Hyperparameters are chosen on the validation set.

\begin{table*}[t!]
\begin{center}
\small
\begin{tabular}{llll|c|c|c|cc}

Model             & Combiner & Text Encoders & Image Encoder& Turn 1 & Turn 2 & Turn 3 & \multicolumn{2}{c}{All} \\
                  &                &   R@1  & R@1    & R@1    & R@1 & R@1    & R@1 &  R@5 \\
\hline
IR Baseline       & n/a & n/a & n/a            & -      &  -     & -      & 2.15 & 5.86\\
\hline
{\sc TransResNet}$_{RET}$  & MM-Att & Separate  & ResNet152      & 35.7 & 44.5	& 40.5 & 40.2 & 67.0\\
{\sc TransResNet}$_{RET}$  & MM-Sum & Separate  & ResNet152      & 34.5 & 46.0 & 41.3 & 40.6 & 67.2\\
\hline
{\sc TransResNet}$_{RET}$ & MM-Sum & Shared   & ResNeXt-IG-3.5B & 53.6 &	47.0 & 41.3 & 47.3 & 73.1\\
{\sc TransResNet}$_{RET}$ & MM-Att & Shared   & ResNeXt-IG-3.5B & \textbf{54.4} & 49.0 & 43.3	& 48.9 & 74.2\\
{\sc TransResNet}$_{RET}$ & MM-Att & Separate & ResNeXt-IG-3.5B & 53.5 & 50.5 & 43.8	& 49.3 & 74.7\\
{\sc TransResNet}$_{RET}$ & MM-Sum & Separate & ResNeXt-IG-3.5B & 54.0 &	\textbf{51.9} & \textbf{44.8}	& \textbf{50.3} & \textbf{75.4}\\
\end{tabular}
\end{center}
\caption{Module choices on \textsc{\imagechat}.
We compare different module variations for {\sc TransResNet}$_{RET}$.
\label{table:imagechat_results}
}
\end{table*}

\begin{table*}[htp]
\begin{center}
\small
\begin{tabular}{l|cccc|cccc}
          & \multicolumn{4}{c}{ {\sc TransResNet}$_{RET}$ (R@1/100 )}  &
          \multicolumn{4}{c}{ {\sc TransResNet}$_{GEN}$  (ROUGE-L)} \\
Modules   &   Turn 1 & Turn 2 & Turn 3 &   All  &   Turn 1 & Turn 2 & Turn 3 &   All\\ 
\hline
Image Only & 37.6 &	28.1 & 20.7	& 28.7 & 21.1 & 21.9 & 22.4 & 21.8\\
Style Only & 18.3 & 15.3 & 17.0 & 16.9 & 20.2 & 20.9 & 22.0 & 21.0\\
Dialogue History Only & 1.0 &	33.7 & 32.3	& 22.3 & 18.9 & 22.7 & 23.7 & 21.8\\
\hline
Style + Dialogue {\em(no image)} & 18.3 & 45.4 & 43.1	& 35.4 & 20.4 & 24.1 & 24.8 & 23.1 \\
Image + Dialogue {\em(no style)} & 37.6 & 39.4 & 32.6 & 36.5 & 21.3 & 22.8 & 23.6 & 22.6 \\
Image + Style {\em(no dialogue)} & \textbf{54.0} & 41.1	& 35.2 & 43.4 & \textbf{23.7} & 23.2 & 23.8 & 23.5 \\
\hline
Style + Dialogue + Image {\em(full model)} &\textbf{54.0}  & \textbf{51.9} & \textbf{44.8} & \textbf{50.3} & \textbf{23.7} & \textbf{24.2} & \textbf{24.9} & \textbf{24.3}\\
\end{tabular}
\end{center}
\caption{Ablations on \textsc{\imagechat}.
We compare variants of our best {\sc TransResNet} generative and retrieval models (ResNeXt-IG-3.5B image encoder, and MM-Sum + separate text encoders for retrieval) 
where we remove modalities: image, dialogue history and style conditioning, reporting R@1/100 for retrieval and ROUGE-L for generation  for dialogue turns 1, 2 and 3 independently, as well as the average over all turns.
\label{table:ablation_results}
}
\end{table*}

\section{Experiments}

We test our models on the \textsc{\imagechat} and IGC datasets using automatic metrics and human evaluations. We analyze the performance of the different module and architecture choices, as well as ablation studies to determine the importance of each of the model's inputs.

\subsection{Automatic Evaluation on \textsc{\imagechat}}
\paragraph{Module Choices}
We first compare various module configurations of our {\sc TransResNet}$_{RET}$ model, and additionally show the results for a simple information retrieval baseline, in which the candidates are ranked according to their weighted word overlap to the input message.  We measure recall at 1 and 5 (R@1/100 and R@5/100) retrieval metrics, where for each sample there are 100 candidates to rank: 99 random candidates chosen from the test set, and the true label. Note that in human evaluations we  use all the train set candidates.

The results are shown in Table \ref{table:imagechat_results}. 
We report the average metrics for the total task, as well as the breakdown of the performance on each turn of dialogue (turns 1, 2 and 3). The average metrics indicate that using the ResNeXt-IG-3.5B~image encoder features improves performance significantly across the whole task, as we obtain 50.3\% R@1 for our best 
ResNeXt-IG-3.5B~model and only 40.6\% for our best ResNet152 model. When broken down by turn, it appears that the ResNeXt-IG-3.5B~features are particularly important in the first round of dialogue, in which only the image and style are considered, as the difference between their best models increases from 9.7\% in the full task to 19.5\% in the first turn. Our baseline multimodal sum combiner (MM-Sum) outperforms the more sophisticated self-attention (MM-Att) combiner, 
with the latter scoring 49.3\% on the full task. 
Having separate candidate and dialogue history text encoders also works better than sharing weights.

In subsequent experiments we use the best performing system 
for our retrieval model.
As ResNeXt-IG-3.5B~performs best we use that for our generative model going forward as well.

\paragraph{Full \& Ablation Study} 
We now perform experiments for both retrieval and generative models for the full system, and additionally
 we remove modalities (image, style, and dialogue history). 
 For the generative models we report the ROUGE-L metric.
The results are shown in Table \ref{table:ablation_results}, which we now analyze.

{\em{Turn 1:}} In the first round of dialogue the models produce  utterances given the image and style only, as there is no dialogue history yet.
For both models, image is more important than style, but using both together helps.

{\em{Turn 2:}} In the second turn, in which a model produces a response to a first utterance, the models perform similarly when using only the image or only the dialogue history, while performing poorly with just the style. Any combination of two modalities improves the results, with the style + dialogue combination performing slightly higher than the other two. Using all modalities works best.

{\em{Turn 3:}} By the third turn of dialogue, the conversation history proves to be by far the most important in isolation compared to the other two modalities in isolation.  Conditioning on the style+dialogue is the most effective of any combination of two modalities. Again, using all modalities still proves best.

\subsection{Human Evaluations on \textsc{\imagechat}}
\label{imagechat_humanevals}

We  test our final models using human evaluation.

\paragraph{Evaluation Setup} We use a set of 500 images from YFCC-100M that are not present in \textsc{\imagechat} to build a set of three-round dialogues pairing humans with models in conversation. We then conduct evaluations at each round of dialogue for each example in the evaluation set; we have a separate set of human evaluators look at the provided conversation turns, and ask them to compare two possible utterances for the next turn of conversation, given the image, dialogue history and relevant style (which is the same for both human author and model, so there is no advantage). We ask the evaluators in a blind test to choose the ``more engaging'' of the two possible utterances: one from a human, and the other from a model.

\paragraph{Human annotation vs. {\sc TransResNet} model} We compare human-authored utterances to those produced by our models. The human conversations are collected in the same fashion as in \textsc{\imagechat} but on test images. As for humans, the model outputs are conditioned on the image, style and previous dialogue history.
{\sc TransResNet}$_{GEN}$ simply generates a response, whereas
{\sc TransResNet}$_{RET}$  retrieves candidate utterances from the \textsc{\imagechat} training set. The latter is given a separate set of candidates corresponding to the round of dialogue -- e.g. when producing a response to turn 1, the model retrieves from all possible round 1 utterances from the train set (in that case
186,858 possible choices). 

\begin{figure*}[!]
\scriptsize
\begin{small}
\resizebox{\textwidth}{!}{
\begin{tabular*}{\textwidth}{ccl}
\hline
\small{Image} & \small{~~Style} & \small{Conversation}  \\  \hline
\hline
Turn 1 examples\\
\hline
\\[-1.8ex]
\multirow{4}{*}{\includegraphics[height=12.5ex, width=19ex]{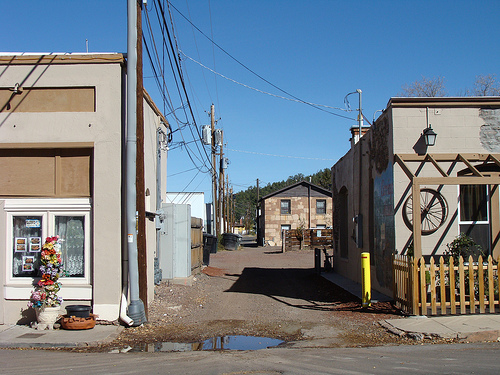}}
\\[-1.5ex]
& & {\em{\small{ Model predictions:}}}  \\
& \small{A: Artful} & \small{This looks like a painting. Something out of a Norman Rockwell.}  \\
& \small{A: Monstrous} &  \small{Some wretched town!}.\\
& \small{A: Earnest} & \multirow{2}{0.55\textwidth}{\small{Yeah, we have finally arrived at Grandpa's old barnhouse!  I can't wait to get started with the restoration! }}  \\[2.4ex] 
\hline
\\[-1.8ex]
\multirow{4}{*}{\includegraphics[height=12.5ex, width=19ex]{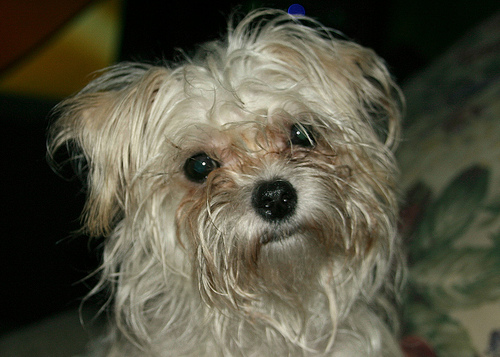}}
\\[-1.5ex]
& & {\em{\small{Model predictions:}}}  \\
& \small{A: Wishful} & \small{I hope one day to have a dog this majestic.}\\
& \small{A: Opinionated} & \small{This puppy looks cold get him a blanket.}  \\
& \small{A: Imaginative} & \small{Puppies are just the universe's way of telling us everything will be okay.}  \\[2.4ex]
\hline
\\[-1.8ex]
\multirow{4}{*}{\includegraphics[height=14.5ex, width=19ex]{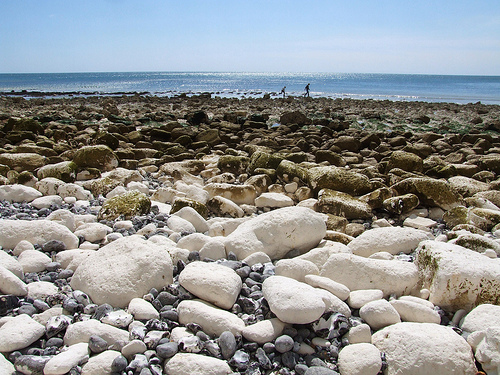}}
\\[-1.5ex]
& & {\em{\small{Model predictions:}}}  \\
& \small{A: Respectful} & \multirow{2}{0.55\textwidth}{\small{What an honor to have beautiful places like these to contemplate natures rocks at their best.}}  \\[2.4ex] 
& \small{A: Shy} & \multirow{2}{0.55\textwidth}{\small{While the area looks nice, I'd stay away, fear of tsunamis leaves me paralyzed at the beach.}}  \\[2.6ex]  
& \small{A: Confused} & {\small{I don't understand why would you put rocks there?}}  \\[0.5ex]
\\[-1.8ex]
\hline
Turn 2 examples\\
\hline
\\[-1.8ex]
\multirow{5}{*}{\includegraphics[height=12.5ex, width=19ex]{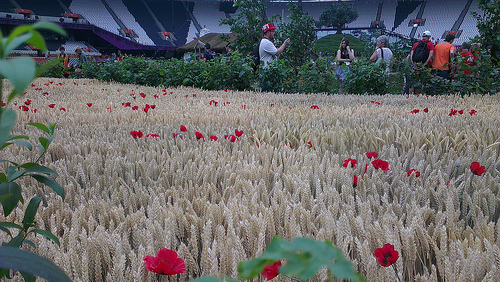}}
\\[-1.5ex]
& \small{A: Empathetic} & \multirow{2}{0.55\textwidth}{\small{I understand that an open field full of beautiful foliage is a testament to our environment.}}  \\[2.4ex] 
& & {\em{\small{Model predictions:}}}  \\
& \small{B: Aloof} & \small{It's just a regular old field.} \\
& \small{B: Frivolous} & \small{So is every grass in the country}\\[1.4ex]
\hline
\\[-1.8ex]
\multirow{4}{*}{\includegraphics[height=12.5ex, width=19ex]{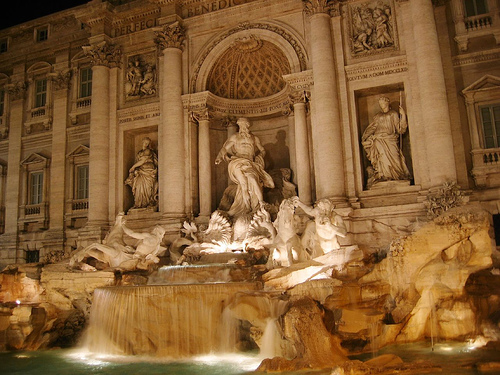}}
\\[-1.5ex]
& \small{A: Playful} & \small{i would jump inside that waterfall}  \\
& & {\em{\small{Model predictions:}}}  \\
& \small{B: Questioning} & \small{Do you really think it is feasible?}  \\
& \small{B: Amusing} & \small{Well I would try to top you with a backflip.}\\[2.4ex]
\hline
\\[-1.8ex]
\multirow{4}{*}{\includegraphics[height=12.5ex, width=19ex]{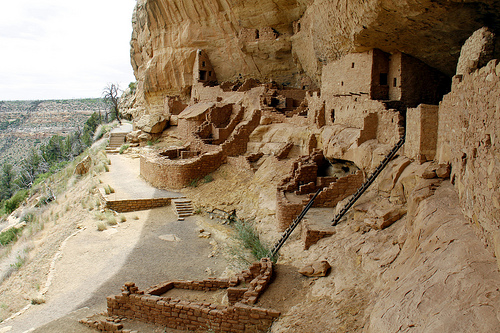}}
\\[-1.5ex]
& \small{A: Maternal} & \small{It must be very dangerous if children play there.}  \\
& & {\em{\small{Model predictions:}}}  \\
& \small{B: Whimsical} & \small{Nonsense, this would be a great place to play!}  \\
& \small{B: Blunt} & \small{I would rather find some lawn. Those rocks look uncomfortable}\\[2.4ex]
\hline
Turn 3 examples\\
\hline
\\[-1.8ex]
\multirow{4}{*}{\includegraphics[height=12.5ex, width=19ex]{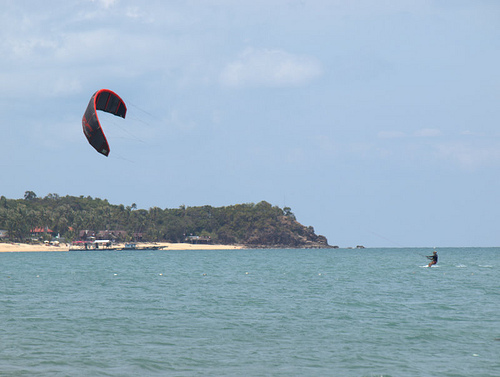}}
\\[-1.5ex]
& \small{A: Happy} & \small{This looks like something fun and exciting to do on vacation!} \\
& \small{B: Anxious} & \small{oh dear what happens if he gets taken with the wind}  \\
& & {\em{\small{Model prediction:}}}  \\
& \small{A: Happy} & \small{Ooh, the wind looks perfect to fly this awesome kite!} \\[2.4ex] 
\hline
\\[-1.8ex]
\multirow{4}{*}{\includegraphics[height=12.5ex, width=19ex]{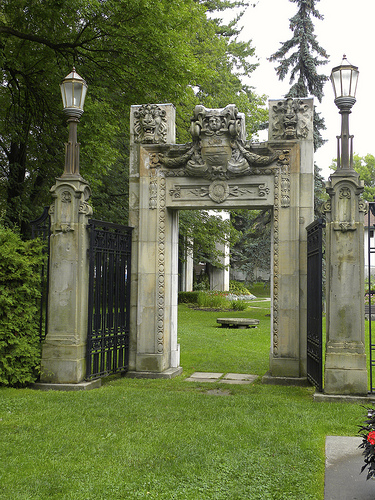}}
\\[-1.5ex]
& \small{A: Pompous} & \multirow{2}{0.55\textwidth}{\small{This archway is so dull and boring. I could have done a much better job with the garden in my own back yard.}}  \\[2.4ex]
& \small{B: Offhand} & \small{Its about as boring as you!!}  \\
& & {\em{\small{Model prediction:}}}  \\
& \small{A: Pompous} & \small{Please. I could show you much prettier places in my own back yard.} \\[2.4ex] 
\hline
\\[-1.8ex]
\multirow{4}{*}{\includegraphics[height=14.5ex, width=19ex]{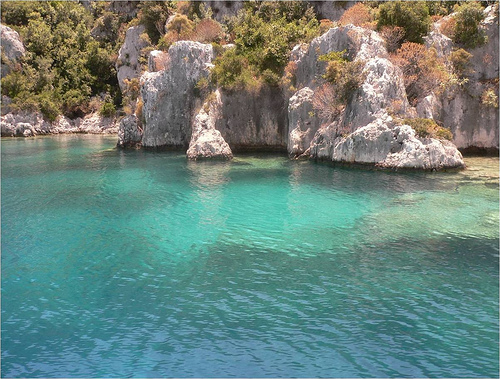}}
\\[-1.5ex]
& \small{A: Sweet} & \multirow{2}{0.55\textwidth}{\small{This is so beautiful, I love nature and would love to see this place and meet the people who live there!}}  \\[2.4ex]
& \small{B: Spontaneous} & \small{Then lets go there right now! Just you and me.}  \\
& & {\em{\small{Model prediction:}}}  \\
& \small{A: Sweet} & \small{I wish you could take me there, I would love to go with you!} \\[2.4ex]
\hline
\end{tabular*}
}
\end{small}
\caption{Example predictions from our {\sc TransResNet$_{RET}$} {\small{(MM-Sum)}} model on the evaluation set using all candidates for turns 1--3 . Two speakers A \& B with given style traits discuss a photo. The dialogue context before the model prediction is completed by humans, followed by one or more possible model responses, given different style conditioning. 
The model clearly uses the image, given style and dialogue history in formulating its response.
}
\label{evalsetimages_turn3}
\end{figure*}

\begin{figure}[t!]
\begin{center}
\includegraphics[trim={0 0 0 3mm}, clip, width=0.48\textwidth]{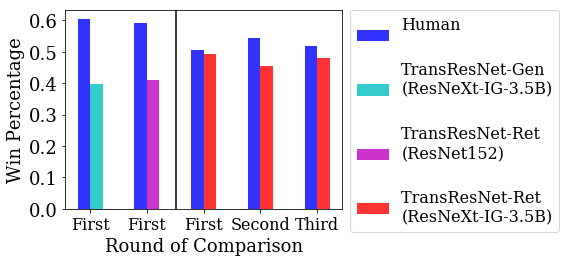}
\caption{Human evaluations on \textsc{\imagechat}.
Engagingness win rates of pairwise comparisons between human utterances and
{\sc TransResNet}$_{RET}$ (ResNet152 or ResNeXt-IG-3.5B) or {\sc TransResNet}$_{GEN}$,
comparing over the rounds of dialogue.
}
\label{fig:humanevals1}
\end{center}
\end{figure}

The results are shown in Fig. \ref{fig:humanevals1}, comparing all models on the first round (left): {\sc TransResNet}$_{GEN}$ and  {\sc TransResNet$_{RET}$}
using ResNeXt-IG-3.5B,  and  {\sc TransResNet$_{RET}$} using  ResNet152 features.
As in automatic evaluations, ResNet152 features performed more poorly.
The retrieval model outperformed the generative model, a result 
that has been observed in other (text-only) dialogue tasks \citep{dinan2019wizard,zhang2018personalizing}.
In turn 1,  {\sc TransResNet$_{RET}$} (ResNeXt-IG-3.5B)  has a win rate against humans of 49.4\% (difference not significant using a binomial two-tailed test, $p>0.5$),
while both other models are significantly outperformed by humans ($p<2 \times 10^{-7}$ compared to ResNet152 features), showing the importance of our retrieval architecture and image feature choices. 
We thus compare only  {\sc TransResNet$_{RET}$}
 (ResNeXt-IG-3.5B) to humans in all three turns (Fig.~\ref{fig:humanevals1}, right).
That model performs well, with  an overall win rate against humans of 47.7\% (difference is significant, $p<7\times10^{-5}$).
Example predictions of  {\sc TransResNet}$_{RET}$ (ResNeXt-IG-3.5B) 
are given in 
Figure \ref{evalsetimages_turn3}.

\subsection{Transfer to the IGC Task}
\label{transfer_to_igc_task}

To test the strength of our task and models we consider transfer to the IGC of task of 
\citet{mostafazadeh2017image}. In particular, we focus on their response task, which provides an image and a dialogue history of two utterances: a context utterance, followed by a question. The task is to then produce a response.
This is clearly related to our task, except it focuses on answering questions, which our task does not. Our task is more varied as it was collected  in an unconstrained way, unlike in IGC where they were asked to write a question. Nevertheless, assuming a question contains a {\em ?} or starts with {\em who}, {\em what}, {\em when}, {\em where}, {\em why} or {\em how}, our dataset contains 
40,076 training utterances that are questions (11.3\% of the data) and so it could be possible to produce responses to them. Without any fine-tuning at all, we thus simply took exactly the same best trained
models
and used them for their question response task as well.

Unfortunately, after contacting the authors of \citet{mostafazadeh2017image} they no longer have the predictions of their model available, nor have they made available the code for their human evaluation setup. However, the test set is available. We therefore attempted to reproduce the same setup as in their experiments, which we will also make publicly available upon acceptance. 

\begin{figure}[t]
\begin{center}
\includegraphics[trim={0 0 0 3mm}, clip, width=0.48\textwidth]{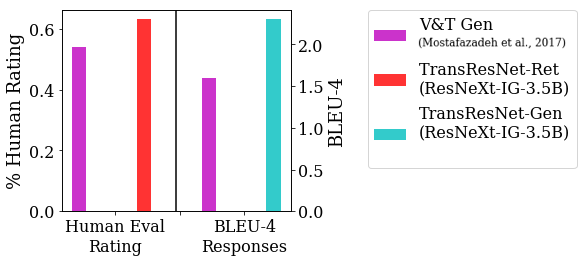}
\caption{IGC Evaluations. The best model from \citet{mostafazadeh2017image} is compared to our best {\sc TransResNet$_{RET}$} and {\sc TrasnResNet$_{GEN}$} models. On the left, annotator's ratings of responses from the models are shown as a percentage of the annotator's ratings of human responses. On the right, BLEU-4 scores on the response task are shown.
}
\label{fig:humanevals2}
\end{center}
\end{figure}

\paragraph{Automatic Evaluation} We measure our best {\sc TransResNet$_{GEN}$} model's performance on the IGC test set in terms of BLEU-4. The results are shown in  Fig.~\ref{fig:humanevals2} (right). We find that our model outperforms the model from \citet{mostafazadeh2017image}, achieving a score of 2.30 compared to 1.49.

\paragraph{Human Evaluation} 
We compare the provided human response (from the test set) with 7 variants of our  \textsc{TransResNet}$_{RET}$
model (mimicking their setup), whereby we have our model condition on 7 styles for which it performed well on evaluations in section \ref{imagechat_humanevals}.  Annotators rated the quality of responses on a scale from 1 to 3, where 3 is the highest, reporting the mean over $\sim$2k questions.
We then scale that by the score of human authored responses, to give a percentage. 
 The results are shown in  Fig.~\ref{fig:humanevals2} (left).
 Our model narrows the gap between human and model performance, yielding a higher percentage of the human score (62.9\% vs. 54.2\%).
More detailed results and example predictions of our model can be found in 
Appendices \ref{igc_appendix} and \ref{igc_appendix2}, including examples of highly rated and poorly rated outputs from our model.

\section{Conclusion}
This paper presents an approach for improving the way machines can generate grounded conversations that humans find engaging. Focusing on the case of chit-chatting about a given image,  a naturally useful application for end-users of social dialogue agents, 
this work shows that our best
proposed  model can generate grounded dialogues that humans prefer over dialogues with other fellow humans almost half of the time (47.7\%). This result is made possible by the creation of a new dataset {\sc Image-Chat}\footnote{\tiny\url{http://parl.ai/projects/image_chat}}. 

Our work shows that we are close to having models that humans can relate to in chit-chat conversations, which could set new ground for social dialogue agents. 
However, our retrieval models outperformed their generative versions; closing that gap is an important challenge for  the community.
While our human evaluations were on short conversations, initial investigations indicate the model as is can extend to longer chats, see Appendix \ref{long_form_chat_example}, which should be studied in future work.
The next challenge will also be to combine this engagingness with other skills, such as world knowledge \citep{antol2015vqa} relation to personal interests \citep{zhang2018personalizing}, and task
proficiency.

\fontsize{9.0pt}{10.0pt} \selectfont
\bibliography{acl.bib}
\bibliographystyle{acl_natbib}

\appendix

\clearpage
\section{More Details of IGC Evaluations}
\normalsize

In this section we describe a few choices we made and implementation details regarding the IGC human evaluation in the section 
regarding Transfer to the IGC Task.

\paragraph{Multiple Traits} In the IGC human evaluation setup from \cite{mostafazadeh2017image}, human annotators were shown eight choices when rating the quality of responses to questions: seven responses from various models, and one human response. To mirror this setup as closely as possible, we chose seven of our highest performing style traits to condition on to display in addition to the human response. We show the results of each trait in Table \ref{table:igc_full_evals}. 

\paragraph{Automatic Evaluation} In \cite{mostafazadeh2017image}, the authors provide BLEU scores for their models in an attempt to evaluate their effectiveness via automated metrics. The authors note that the scores are very low, ``as is characteristic for tasks with intrinsically diverse outputs." Additionally, it has been shown in \cite{Shuster_2019_CVPR} that BLEU scores for image captioning retrieval models are generally far lower than those of generative models (as retrieval models do not optimize for such a metric), and yet human evaluations can show the complete opposite results.  In fact, in that work retrieval models were shown to be superior to generative models in human evaluations, which is why we adopted them here.
For these reasons we omit BLEU scores of our retrieval models on the IGC test set as uninteresting. We do however compare BLEU scores with our generative model in the main paper.

\paragraph{Test Set Size} The IGC test set provides the urls to all 2591 images for which (context, question, response) tuples were collected. We were only able to recover 2195 images from this initial set, as some of the urls provided are no longer associated with the corresponding images. Thus, our human evaluations are conducted on this subset.

\begin{table}[]
\begin{center}
\small
\begin{tabular}{l|c}
Style & Score \\
\hline
Neutral  & 1.55 \\
Charming & 1.55 \\
Extravagant & 1.55 \\
Calm & 1.57 \\ 
Sweet & 1.58 \\
Spirited & 1.60 \\
Enthusiastic & 1.61 \\
\hline
Human & 2.55
\end{tabular}
\end{center}
\caption{IGC Human Evaluation on responses from our \textsc{TransResNet MM-Sum} model conditioned on various personalities. Responses were rated on a quality scale from 1 to 3, where 3 is the highest. 
\label{table:igc_full_evals}
}
\end{table}

\clearpage
\normalsize
\begin{figure*}[]
\section{\textsc{Image-Chat} Human Annotation Setup}
\label{annotation_setup}
\label{fig:first_response_instructions}
\includegraphics[width=0.99\textwidth]{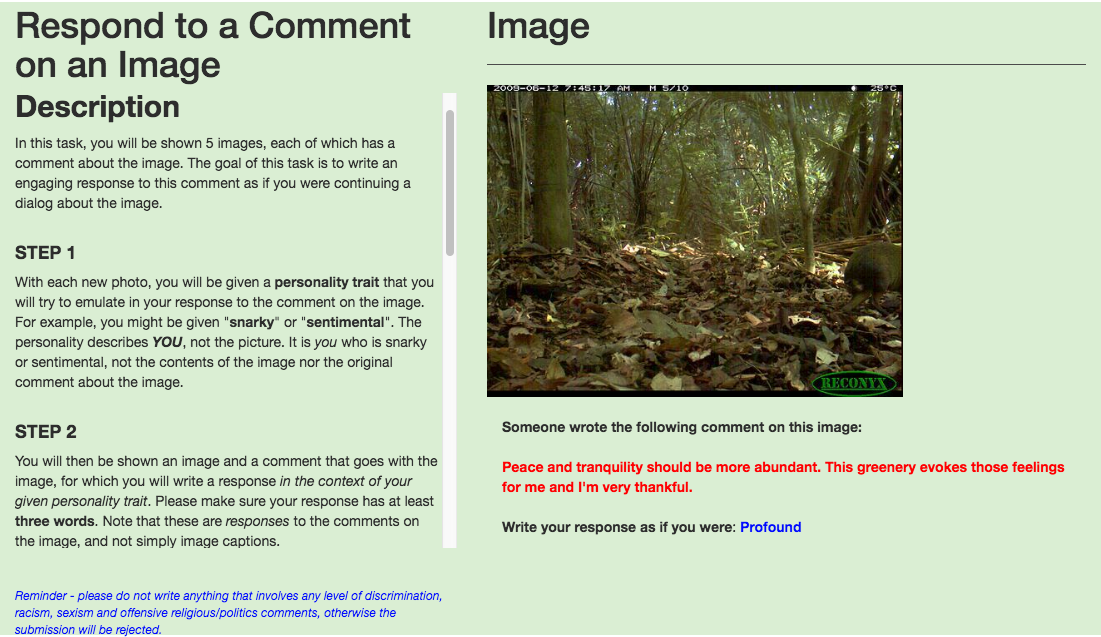}
\caption{Instructions pane for crowdworkers when collecting the second round of dialogue.}
\label{fig:second_response_instructions}
\includegraphics[width=0.99\textwidth]{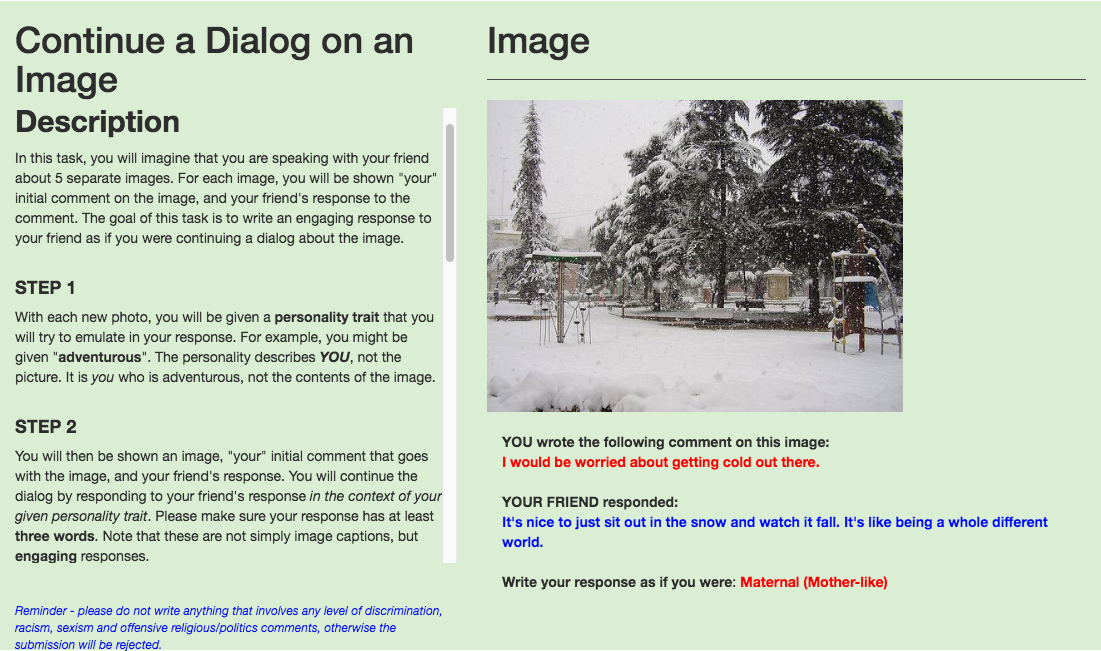}
\caption{Instructions pane for crowdworkers when collecting the third round of dialogue.}
\end{figure*}

\clearpage
\begin{figure}[]
\section{\textsc{Image-Chat} Human Evaluation Setup}
\label{fig:image_chat_human_eval}
\includegraphics[width=0.99\textwidth]{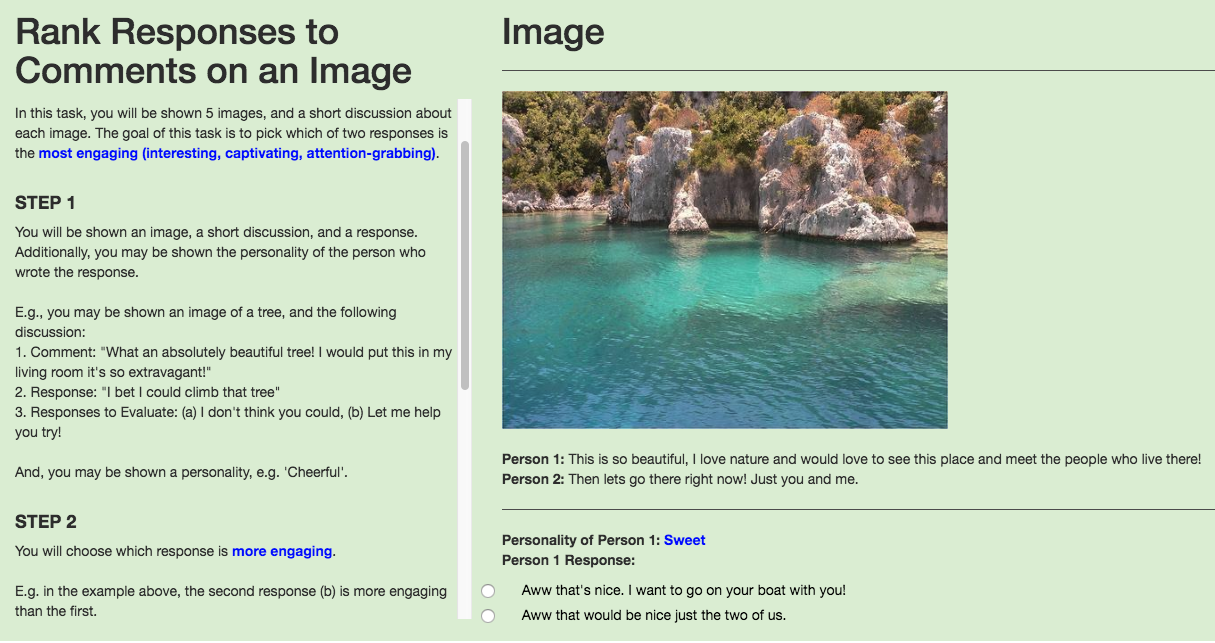}
\caption{Instructions pane for crowdworkers when collecting the \textsc{Image-Chat} Evaluations.}
\end{figure}

\begin{figure}[]
\section{IGC Human Evaluation Setup}
\label{fig:igc_annotation_setup}
\includegraphics[width=0.99\textwidth]{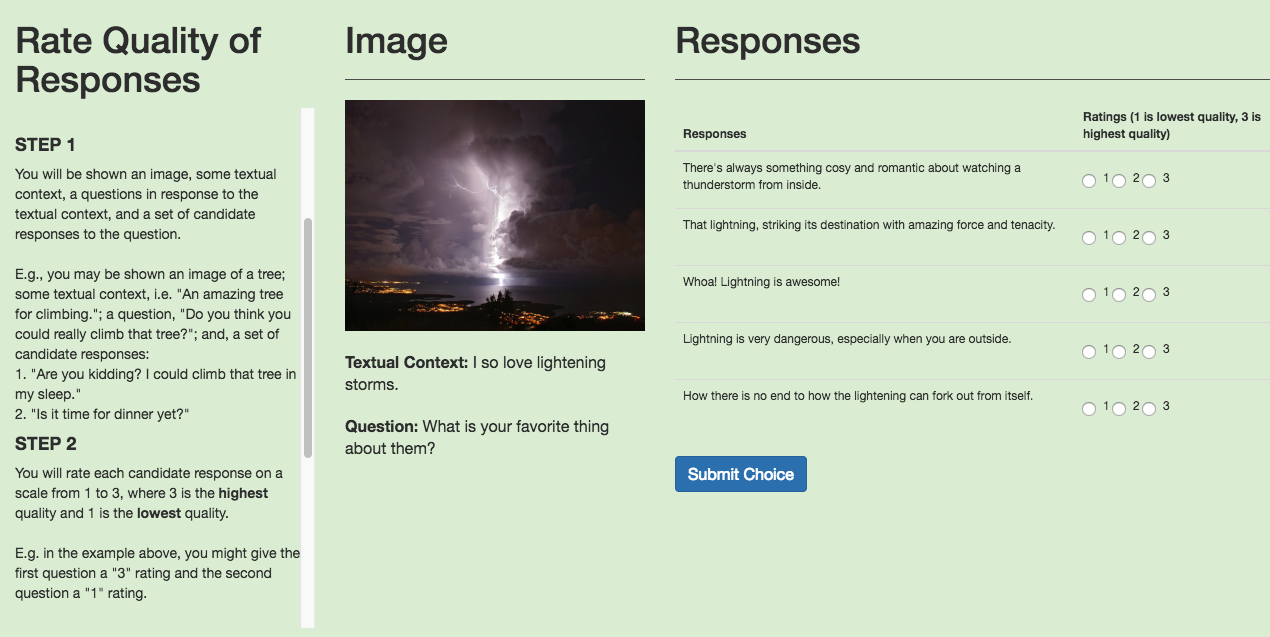}
\caption{Instructions pane for crowdworkers when collecting the IGC Evaluations.}
\end{figure}

\begin{table*}[t]
\section{Highly Rated Examples from IGC}
\label{igc_appendix}

\centering
\begin{small}
\begin{tabular*}{\textwidth}{ccl}
\hline
\small{Image} & \small{IGC Round} & \small{Output}  \\  \hline
\hline
\\[-1.8ex]
\multirow{4}{*}{\includegraphics[height=13ex, width=19ex]{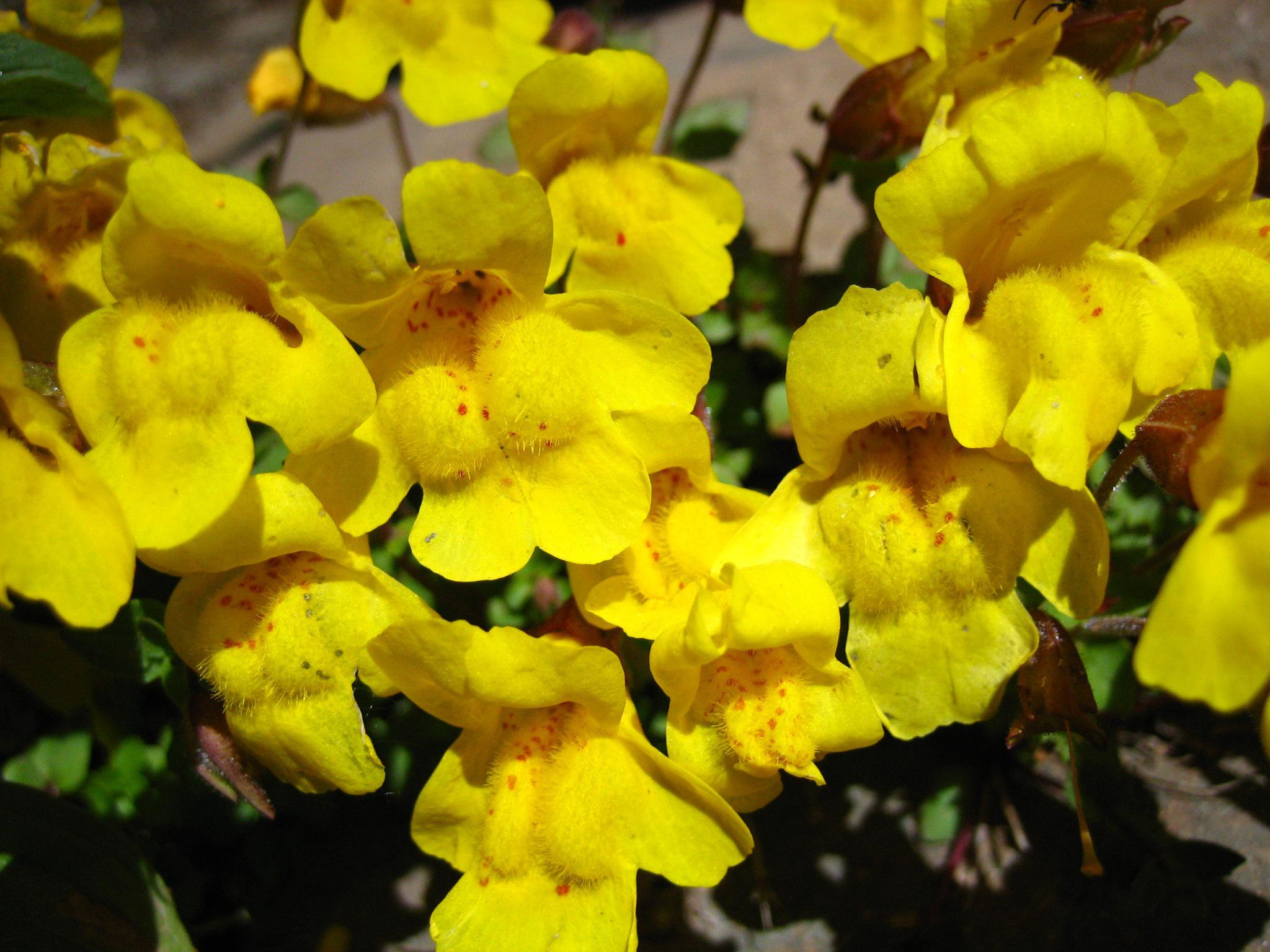}}
\\[-1.5ex]
& \bf{\small{Context}} & \small{These flowers are growing in my back yard.}  \\
& \bf{\small{Question}} & \small{What type of flowers are they?}  \\
& & \\
& \bf{\small{Human Response}} & \small{I don't know but they are so pretty.} \\
& \bf{\small{TransResNet MM-Sum}} & \small{I don't know but these flowers are gorgeous and look so bright!} \\[0.5ex]
\hline
\\[-1.8ex]
\multirow{4}{*}{\includegraphics[height=13ex, width=19ex]{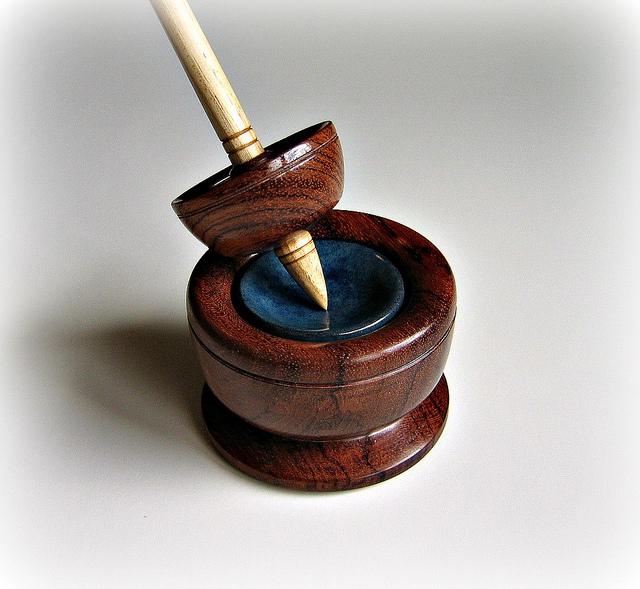}}
\\[-1.5ex]
& \bf{\small{Context}} & \small{I bought this at a flea market.}  \\
& \bf{\small{Question}} & \small{What is this for?}  \\
& & \\
& \bf{\small{Human Response}} & \small{I think it's a mortar and pestle.} \\
& \bf{\small{TransResNet MM-Sum}} & \small{I'm not sure, but you could sell it for some cash!} \\[0.5ex]
\hline
\\[-1.8ex]
\multirow{4}{*}{\includegraphics[height=13ex, width=19ex]{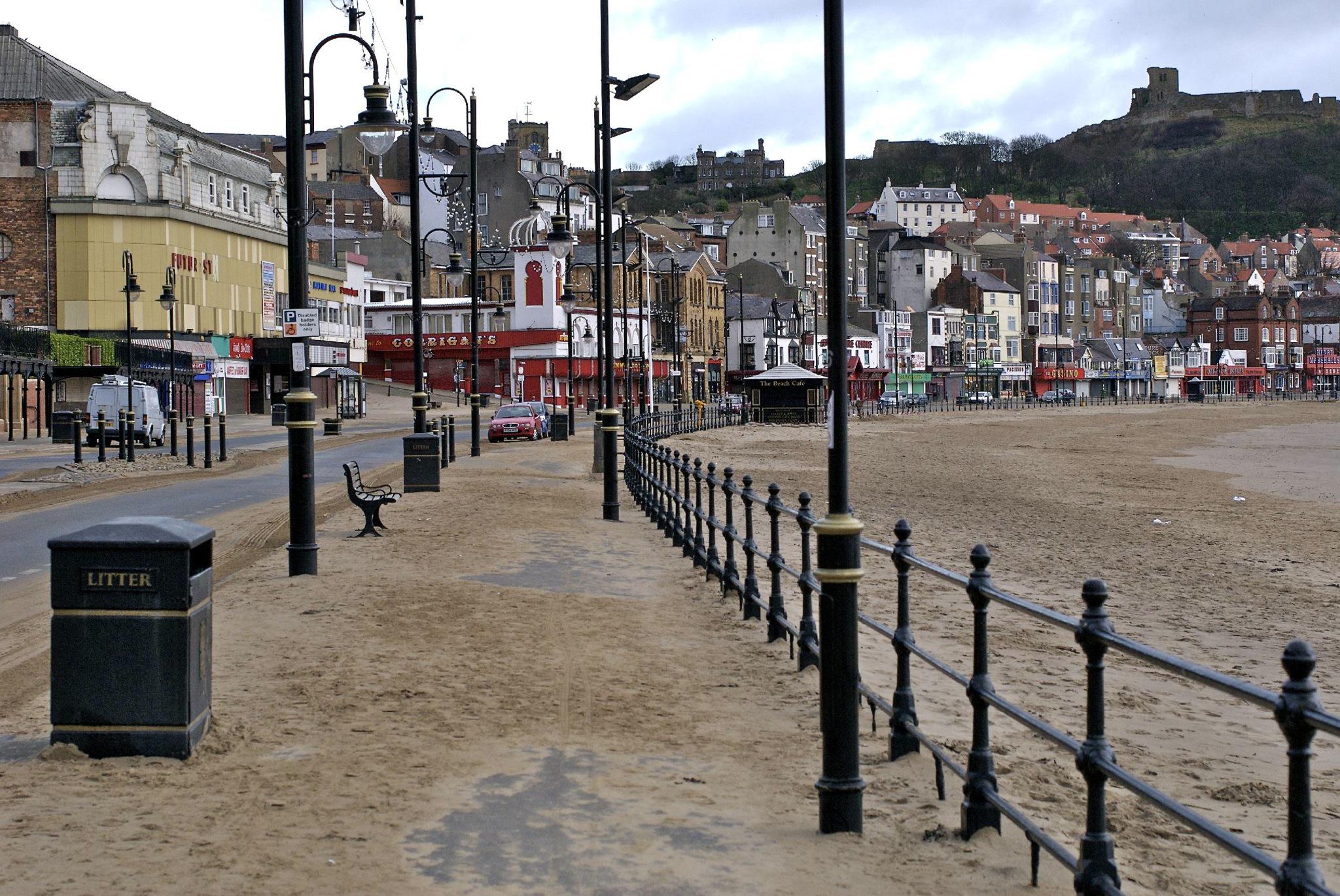}}
\\[-1.5ex]
& \bf{\small{Context}} & \small{I'd love to visit this city.}  \\
& \bf{\small{Question}} & \small{Is this some city in Europe?}  \\
& & \\
& \bf{\small{Human Response}} & \small{It definitely looks like it.} \\
& \bf{\small{TransResNet MM-Sum}} & \small{Oh my goodness, yes! I love Amsterdam!} \\[0.5ex]
\hline
\\[-1.8ex]
\multirow{4}{*}{\includegraphics[height=13ex, width=19ex]{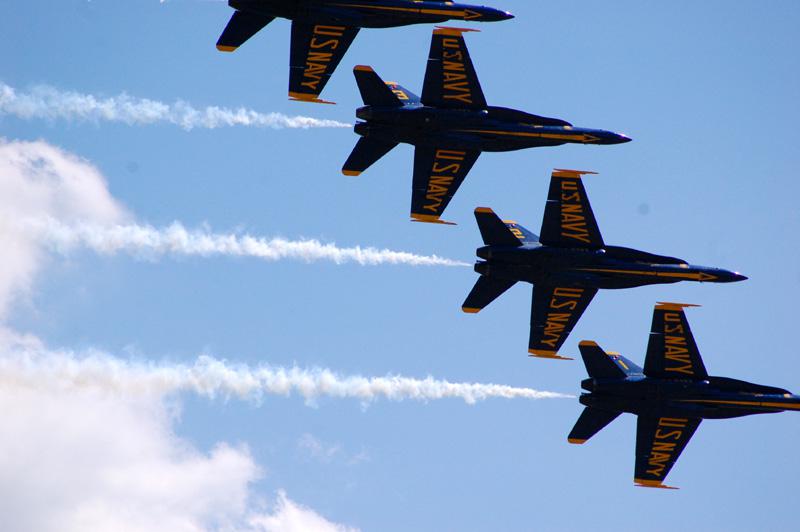}}
\\[-1.5ex]
& \bf{\small{Context}} & \small{These just flew over my house!}  \\
& \bf{\small{Question}} & \small{Wow, aren't those Blue Angels?}  \\
& & \\
& \bf{\small{Human Response}} & \small{Yeah, they practice in this area for their shows.} \\
& \bf{\small{TransResNet MM-Sum}} & \small{These are the Blue Angels, they are truly one of a kind} \\[0.5ex]
\hline
\\[-1.8ex]
\multirow{4}{*}{\includegraphics[height=13ex, width=19ex]{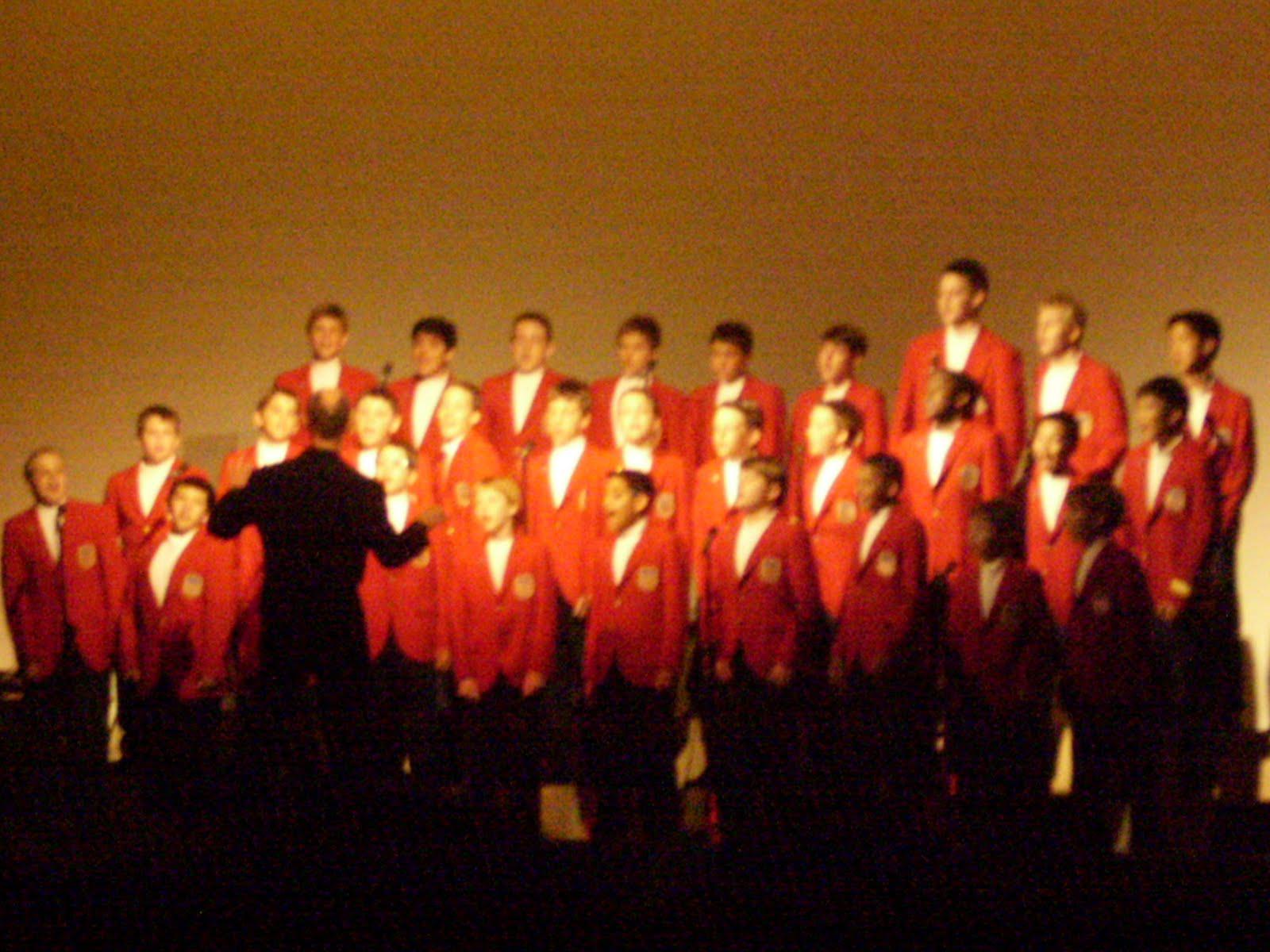}}
\\[-1.5ex]
& \bf{\small{Context}} & \small{My nephew's choir did so well yesterday!}  \\
& \bf{\small{Question}} & \small{That is great.  You must be proud}  \\
& & \\
& \bf{\small{Human Response}} & \small{I am so proud,  because he's been working really hard.} \\
& \bf{\small{TransResNet MM-Sum}} & \small{I am so proud of all that they accomplished.} \\[0.5ex]
\hline
\\[-1.8ex]
\multirow{4}{*}{\includegraphics[height=13ex, width=19ex]{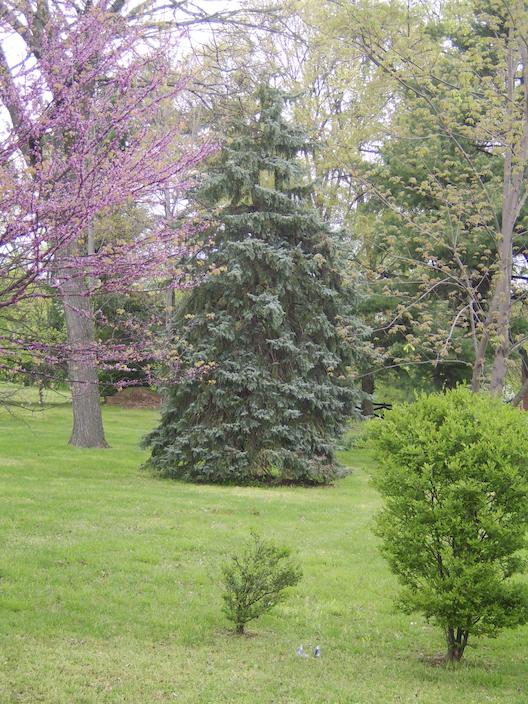}}
\\[-1.5ex]
& \bf{\small{Context}} & \small{This would be a great place to have a picnic.}  \\
& \bf{\small{Question}} & \small{yes it would it seems very relaxing doesnt it?}  \\
& & \\
& \bf{\small{Human Response}} & \small{Yes, and I would quite enjoy spending a day here.} \\
& \bf{\small{TransResNet MM-Sum}} & \small{I think a picnic is more than a great possibility!} \\[0.5ex]
\hline
\end{tabular*}
\end{small}
\caption{Highly rated examples from the IGC dataset test split where {\sc TransResNet$_{RET}$} MM-Sum responses were rated the highest (score of 3) by human evaluators.}
\label{igc_examples_table}
\end{table*}

\begin{table*}[t]
\section{Poorly Rated Examples from IGC}
\label{igc_appendix2}
\centering
\begin{small}
\begin{tabular*}{\textwidth}{ccl}
\hline
\small{Image} & \small{IGC Round} & \small{Output}  \\  \hline
\hline
\\[-1.8ex]
\multirow{4}{*}{\includegraphics[height=13ex, width=19ex]{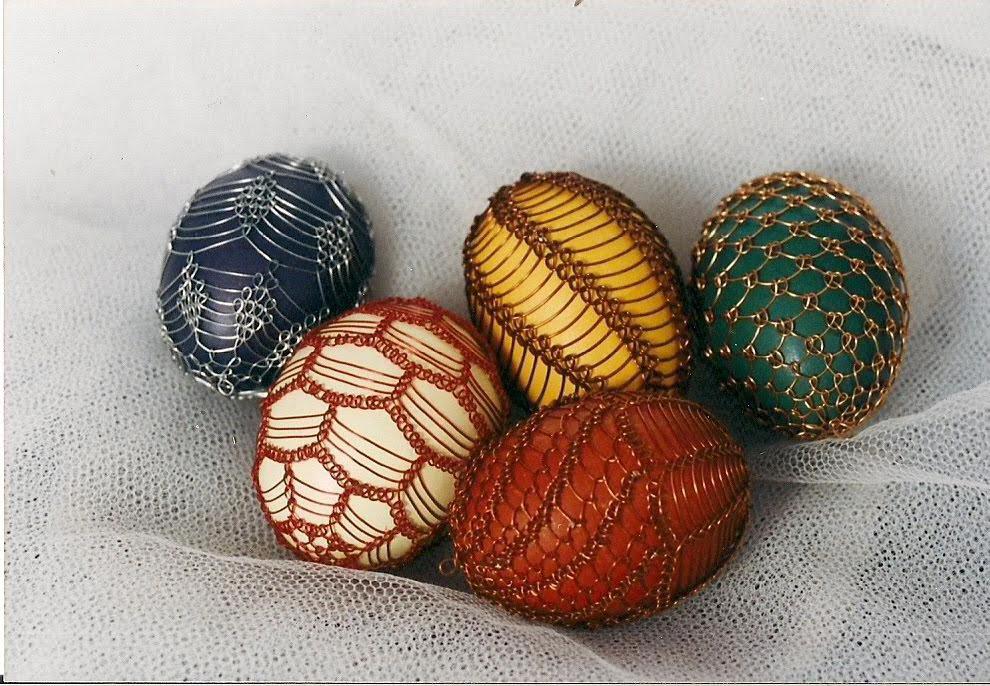}}
\\[-1.5ex]
& \bf{\small{Context}} & \small{These are Easter eggs we made for art class.}  \\
& \bf{\small{Question}} & \small{How did you make them?}  \\
& & \\
& \bf{\small{Human Response}} & \small{We started with a glass base and went from there.} \\
& \bf{\small{TransResNet MM-Sum}} & \small{OHMYGOSH EASTER EGG HUNT! I want to hunt for Easter eggs too!} \\[0.5ex]
\hline
\\[-1.8ex]
\multirow{4}{*}{\includegraphics[height=13ex, width=19ex]{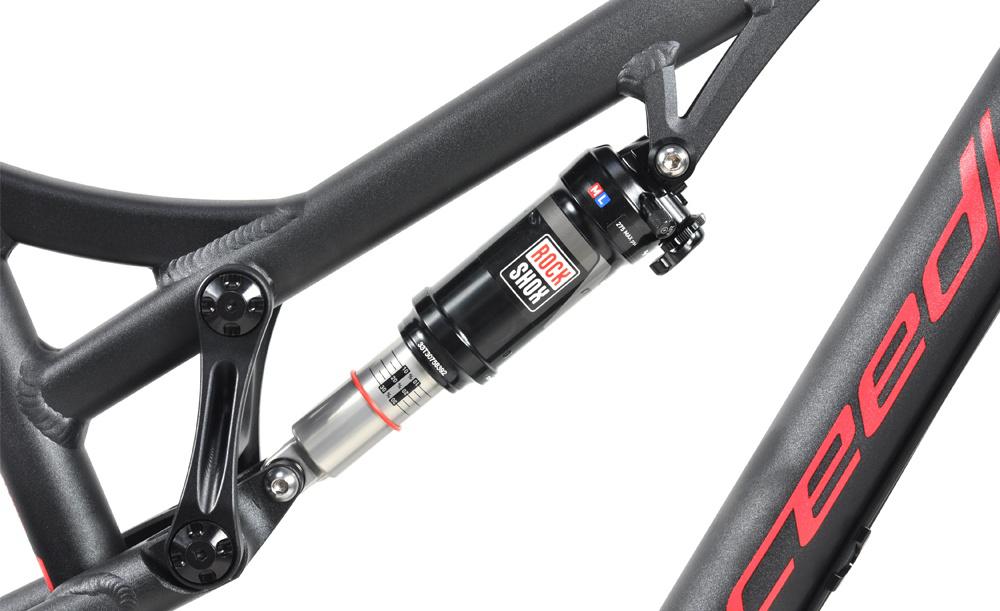}}
\\[-1.5ex]
& \bf{\small{Context}} & \small{I'm thinking of getting these shocks on my bike.}  \\
& \bf{\small{Question}} & \small{Are the shocks black in color?}  \\
& & \\
& \bf{\small{Human Response}} & \small{Yes, except for the silver portion at the end.} \\
& \bf{\small{TransResNet MM-Sum}} & \small{Just got my new wheels!} \\[0.5ex]
\hline
\\[-1.8ex]
\multirow{4}{*}{\includegraphics[height=13ex, width=19ex]{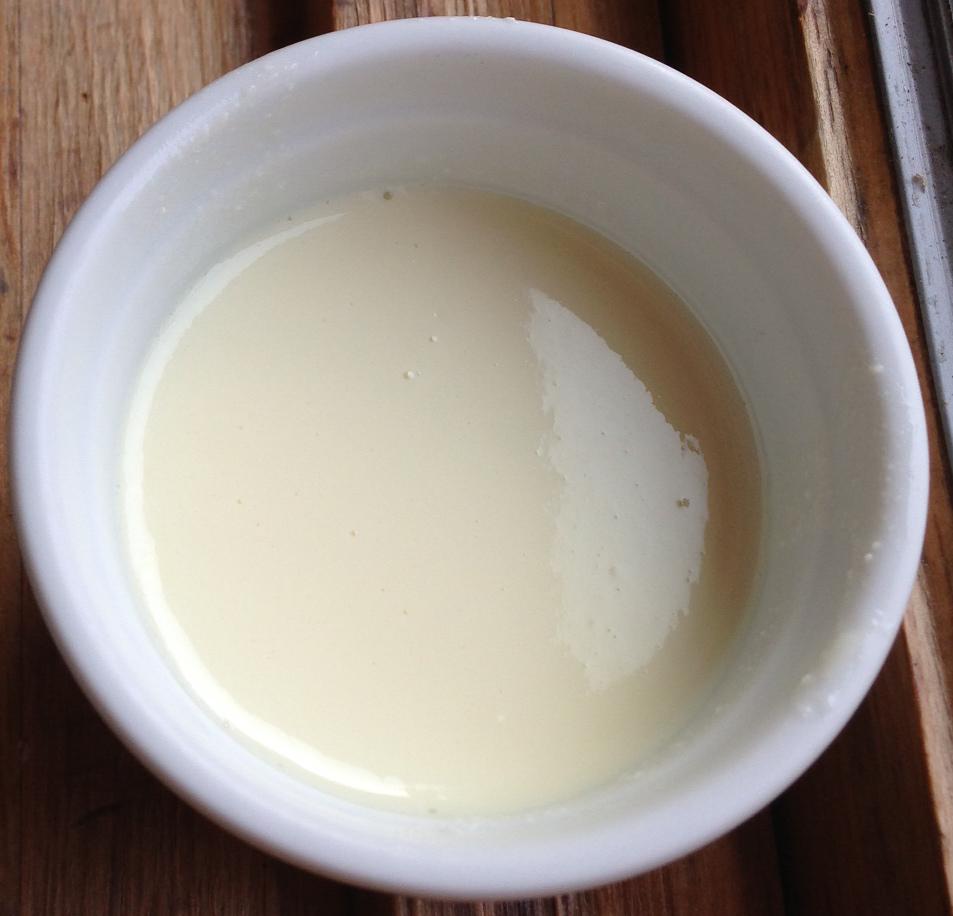}}
\\[-1.5ex]
& \bf{\small{Context}} & \small{I cannot decide if this is milk or something else.}  \\
& \bf{\small{Question}} & \small{It looks like milk, but it could also be kefir.}  \\
& & \\
& \bf{\small{Human Response}} & \small{What is kefir?} \\
& \bf{\small{TransResNet MM-Sum}} & \small{You can get all the protein you want form the sugar.} \\[0.5ex]
\hline
\\[-1.8ex]
\multirow{4}{*}{\includegraphics[height=13ex, width=19ex]{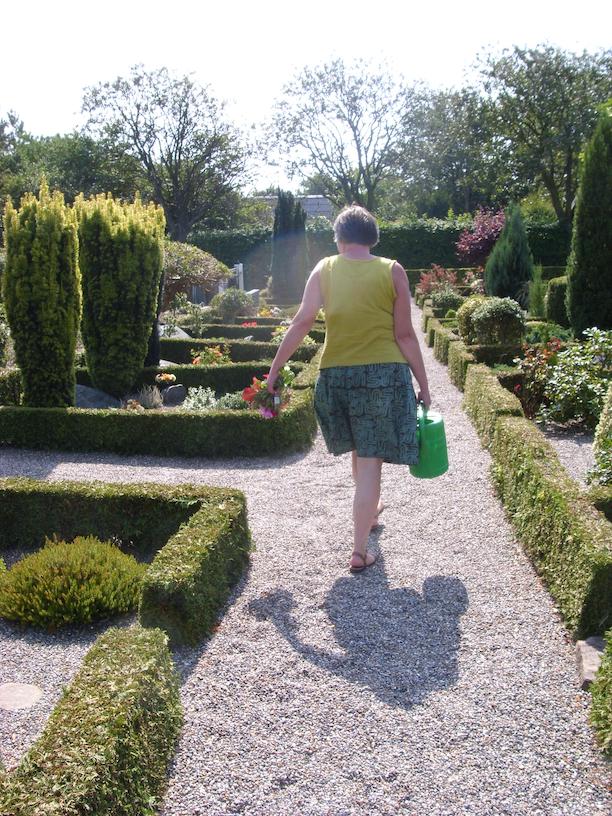}}
\\[-1.5ex]
& \bf{\small{Context}} & \small{This woman is visiting the grave of a loved one.}  \\
& \bf{\small{Question}} & \small{Where is the cemetery?}  \\
& & \\
& \bf{\small{Human Response}} & \small{The cemetary is at the foot a beautiful quartz mountain.} \\
& \bf{\small{TransResNet MM-Sum}} & \small{Well she's just a walking, talking garden!} \\[0.5ex]
\hline
\end{tabular*}
\end{small}
\caption{Low rated examples from the IGC dataset test split where {\sc TransResNet$_{RET}$} MM-Sum responses were rated the lowest (score of 1) by human evaluators.}
\label{igc_examples_table2}
\end{table*}

\clearpage
\begin{figure*}[b]
\section{Longer Chat Example}
\label{long_form_chat_example}
\centering\setlength{\tabcolsep}{0.5em}
\begin{tabular}{l}
\\[-0.62em]
\includegraphics[width=30em, height=15em]{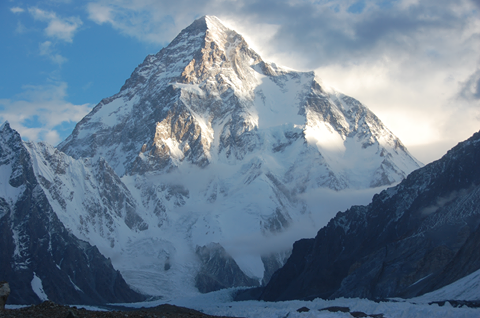} \\
\textit{\small{A: TransResNet$_{RET}$ MM-Sum (Extraordinary)~~~~~~B: Human (no style) }} \\\\
\small{A: It's the most beautiful mountain in the world!}  \\\\
\small{B: It looks really cool! I wonder if you can ski on it}  \\\\
\small{A: They are a mountain range I've climed many times!}  \\\\
\small{B: You have climbed that? That seems really hard}  \\\\
\small{A: I did it; i conquered this climb in only 7 hours and 10 minutes!}  \\\\
\small{B: Wow, that's fast. I would ski down that but I would need a lift to take me up}  \\\\
\small{A: But could you? Could you truly climb this?}  \\\\
\small{B: I really don't think I could}  \\\\
\small{A: Climbing a mountain can give one a special strength, you need to experience it}  \\\\
\small{B: Maybe one day on a smaller mountain}  \\\\
\small{A: It would take hard will and determination to scale that mighty peak}  
\end{tabular}
\caption{Long-form conversation with the model. The model is given a style here, while the human is not.}
\end{figure*}

\begin{table*}[t]
\section{Additional Ablation Results} 
\begin{center}
\small
\begin{tabular}{l|cccc|cccc}
          & \multicolumn{4}{c}{ {\sc TransResNet}$_{GEN}$ (F1)}  &
          \multicolumn{4}{c}{ {\sc TransResNet}$_{GEN}$  (BLEU-4)} \\
Modules   &   Turn 1 & Turn 2 & Turn 3 &   All  &   Turn 1 & Turn 2 & Turn 3 &   All\\ 
\hline
Image Only & 10.8 &	11.0 & 11.2	& 11.0 & 1.1 & 1.3 & 1.2 & 1.2\\
Style Only & 10.4 & 9.8 & 10.4 & 10.2 & 1.4 & 1.5 & 1.4 & 1.4\\
Dialogue History Only & 9.9 & 11.4 & 12.2	& 11.2 & 1.0 & 1.9 & 1.8 & 1.6\\
\hline
Style + Dialogue {\em(no image)} & 9.6 & \textbf{12.5} & \textbf{13.1}	& 11.7 & 1.5 & \textbf{2.1} & \textbf{2.0} & \textbf{1.9} \\
Image + Dialogue {\em(no style)} & 10.7 & 11.1 & 11.7 & 11.2 & 1.1 & 1.7 & 1.6 & 1.5 \\
Image + Style {\em(no dialogue)} & 12.1 & 11.6	& 11.6 & 11.8 & 1.6 & 1.5 & 1.5 & 1.6 \\
\hline
Style + Dialogue + Image {\em(full model)} &\textbf{12.3}  & \textbf{12.5} & \textbf{13.1} & \textbf{12.6} & \textbf{1.7} & \textbf{2.1} & \textbf{2.0} &\textbf{ 1.9}\\
\end{tabular}
\end{center}
\caption{Ablations on \textsc{\imagechat}.
We compare variants of our best {\sc TransResNet} generative model (ResNeXt-IG-3.5B image encoder) 
where we remove modalities: image, dialogue history and style conditioning, reporting F1 and BLEU-4 for generation for dialogue turns 1, 2 and 3 independently, as well as the average over all turns.
\label{table:additional_ablation_results}
}
\end{table*}

\end{document}